\documentclass{article}
\usepackage{spconf,amsmath,amssymb,graphicx,hyperref}
\usepackage{booktabs,multirow,caption}
\title{UniRank: Unified Rank Allocation for Large Language Models}

\name{Chao Han\textsuperscript{1,2}, Yongjie Du\textsuperscript{1}\sthanks{Corresponding author: Yongjie Du (e-mail: yjiedu2023@swu.edu.cn).}, Junjie Tan\textsuperscript{1}, Zihao Xuan\textsuperscript{3}}
\address{\textsuperscript{1}Southwest University, \textsuperscript{2}Northwestern Polytechnical University, \textsuperscript{3}HKUST}
	
\begin{document}

\maketitle

\begin{abstract}
	Low-rank decomposition is a promising compression paradigm for large language models (LLMs), yet its effectiveness hinges on rank budget allocation across weight matrices: uniform or hand-crafted rules ignore module-wise importance, while learning-based allocation incurs substantial training overhead. We formulate rank allocation as a global sorting-and-truncation pipeline that scores every singular component by combining local singular energy with global functional importance, estimated via layer-wise input--output cosine similarity on a tiny calibration set. We show, both geometrically and empirically, that high input--output cosine similarity implies low effective rank. We further propose rank-preserving fine-tuning (RPFT), which adapts only a small subset of retained singular components so that the allocated rank stays bounded without re-decomposition. Experimental results show that UniRank cuts zero-shot perplexity by up to 50\%, improves average reasoning accuracy by 3.0\% over LoRAP at 25\% sparsity, and boosts four SVD-based decomposition methods as a plug-and-play module.
\end{abstract}


\section{Introduction}
\label{sec:intro}

Large language models (LLMs) achieve state-of-the-art performance, yet their
scale incurs prohibitive memory and computational costs for deployment.
Low-rank decomposition has become a mainstream compression paradigm for LLMs,
but its effectiveness hinges on \emph{rank allocation}: how a fixed parameter
budget is distributed across hundreds of weight matrices. Existing strategies
fall into three paradigms. \emph{Uniform} allocation~\cite{wang2025svd}
treats all matrices alike and ignores module-wise importance.
\emph{Heuristic} allocation~\cite{li2024lorap,lu2025flrc} relies on
hand-crafted, architecture-specific rules; module-level schemes such as
MoDeGPT~\cite{lin2025modegpt} spread sparsity across layers via an
input--output cosine-similarity score while truncating all modules of a layer
at the same ratio, and such rules generalize poorly.
\emph{Learning-based} allocation~\cite{qinsi2025dobisvd,gao-etal-2024-adaptive}
optimizes rank selection (e.g., via differentiable masks) at substantial
training cost; gradient- or Hessian-based estimators~\cite{frantar2023sparsegpt}
are likewise too costly at the granularity of singular vectors. No existing
scheme combines singular-component granularity with global, training-free
allocation across the whole model.

We formulate rank allocation as a global \textbf{Sorting-and-Truncation
(S\&T)} pipeline. Every singular triplet $(\sigma_i,\mathbf{u}_i,\mathbf{v}_i)$
of every decomposed weight matrix is scored by the product of two factors: the
normalized singular energy $\sigma_i^2/\|\boldsymbol{\Sigma}\|_F^2$, measuring
how much variance the component preserves within its own matrix, and the
functional importance $\mathcal{I}=1-\mathbb{E}[\cos(\mathbf{x},\mathbf{y})]$
of its host layer, estimated as the input--output cosine similarity on a tiny
calibration set. Geometrically, high cosine similarity means the layer
transformation is nearly zero or nearly collinear with its input---both regimes
of low effective rank---and we corroborate this empirically (Pearson
$r{=}0.8792$). All singular components from all layers are then sorted by this
two-factor score in a single pool and selected in descending order until the
budget is exhausted, with parameter cost accumulated as $m{+}n$ per retained
vector pair of an $m\times n$ matrix. The two factors are complementary: a
layer-level signal alone (as in MoDeGPT) only shifts budgets \emph{between}
layers, whereas component-level ranking further needs the local energy to pick
which components \emph{inside} a layer survive. S\&T is gradient-free---minutes
of calibration---and plugs into any SVD-type decomposition.

Adapting a compressed model brings a second issue: merging LoRA-style
updates~\cite{hu2022lora} into factorized weights forces full-rank
reconstruction, re-decomposition, and re-truncation, causing avoidable
information loss. We therefore introduce \textbf{Rank-Preserving Fine-Tuning}
(RPFT): most retained singular triplets stay frozen, and only a small subset
(8--16 per matrix) is fine-tuned directly in factorized form, so the rank never
exceeds the allocated budget and re-decomposition is eliminated. Combining
S\&T with RPFT yields \textbf{UniRank}. Our contributions are: (i)~a training-free
two-factor allocator that ranks singular components across all matrices under
explicit parameter costs, serving as a plug-and-play module for existing
SVD-type decomposers; (ii)~geometric and empirical evidence grounding
input--output cosine similarity as a proxy for low effective rank;
(iii)~RPFT, which keeps the allocated rank strictly bounded during adaptation;
and (iv)~experiments on Llama-2 and Llama-3 showing up to 50\% lower zero-shot
perplexity,
$+3.0\%$ average reasoning accuracy over LoRAP at 25\% sparsity, and consistent
gains over four decomposition methods.

\begin{figure*}[t]
	\centering
	\includegraphics[width=\textwidth]{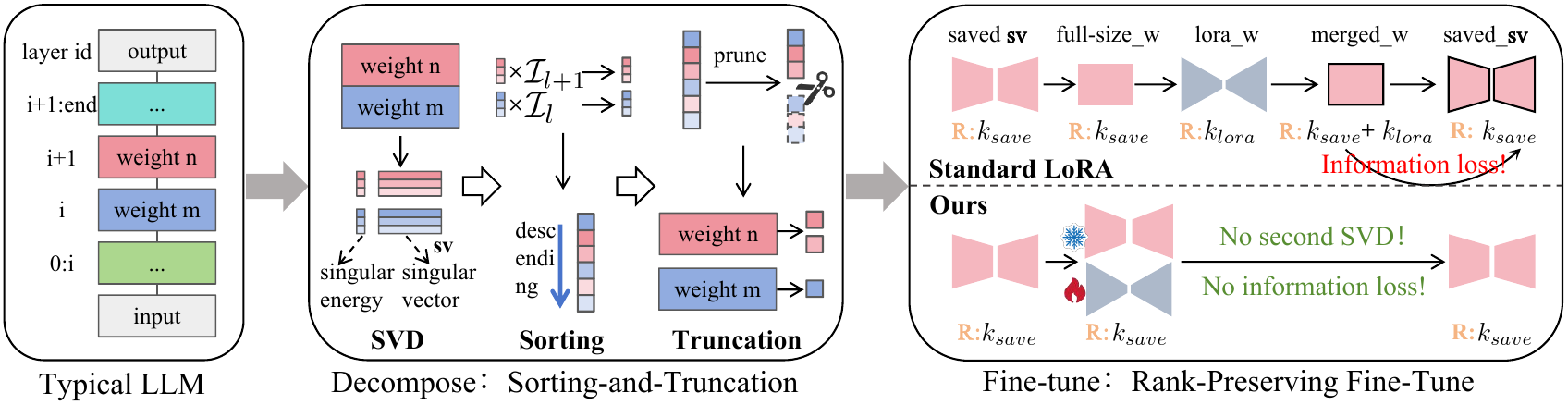}
	\caption{Overview of UniRank. Left: every linear weight is decomposed
	($\mathbf{W}=\sum\sigma_i\mathbf{u}_i\mathbf{v}_i^\top$), and
	Sorting-and-Truncation ranks all singular components across all layers by
	the two-factor score, retaining components until the parameter budget is
	exhausted. Right: Rank-Preserving Fine-Tuning keeps most retained triplets
	frozen (snowflake) and fine-tunes only a small trainable subset (flame)
	within the allocated rank.}
	\label{fig:framework}
\end{figure*}

\section{Method}
\label{sec:method}


Given a weight matrix $\mathbf{W}\in\mathbb{R}^{m\times n}$ of a linear layer,
singular value decomposition (SVD) yields $\mathbf{W}=\sum_{i=1}^{r}\sigma_i
\mathbf{u}_i\mathbf{v}_i^{\top}$ with $\sigma_1\geq\cdots\geq\sigma_r>0$, and a
rank-$k$ approximation keeps the top-$k$ singular triplets. Because one triplet
of an $m\times n$ matrix costs $m{+}n$ parameters, the same rank carries
different parameter costs across matrices, and an effective rank budget must
consider both the importance and the shape of each matrix. UniRank answers
this allocation problem with a global \textbf{Sorting-and-Truncation (S\&T)}
scheme and keeps the resulting budget valid during adaptation through
\textbf{Rank-Preserving Fine-Tuning (RPFT)}; Fig.~\ref{fig:framework}
sketches the pipeline.

\textbf{Global functional importance.} Following the pre-norm residual
structure $f_l(\mathbf{x})=f_{l-1}(\mathbf{x})+\text{Module}_l(\cdot)$ of
modern LLMs~\cite{touvron2023llama2openfoundation,grattafiori2024llama}, we
measure how much a layer transforms its input with the input--output cosine
similarity
\begin{equation}
	\mathcal{I}_l \;=\; 1-\mathbb{E}_{\mathbf{x}\sim\mathcal{D}_c}\!\left[\cos\!\big(f_{l-1}(\mathbf{x}),\,f_l(\mathbf{x})\big)\right],
	\label{eq:importance}
\end{equation}
computed over a tiny calibration set $\mathcal{D}_c$. Cosine similarity has
been used to gauge layer redundancy~\cite{men2024shortgptlayerslargelanguage};
we instead ground it as a proxy for effective rank. The geometric rationale
is illustrated in Fig.~\ref{fig:motivation}. For a residual layer
$\mathbf{y}=\mathbf{x}+f(\mathbf{x})$, a high input--output cosine similarity
admits only two regimes for the residual branch: either $f(\mathbf{x})\approx
\mathbf{0}$, in which case \emph{all} effective singular values of the module
vanish, or $f(\mathbf{x})\approx c\,\mathbf{x}$, in which case the module
degenerates to a nearly isotropic scaling of effective rank one. Both regimes
imply low effective rank, i.e., high compressibility; because scale-invariant
normalization makes the feature \emph{direction} rather than its magnitude
dominate the information flow, this cosine similarity is a well-behaved,
gradient-free proxy for how much rank layer $l$ deserves. We validate the
proxy on Llama-3.2-3B by correlating $\mathcal{I}_l$ with the perplexity
increase caused by per-layer rank truncation at equal retained energy
(Fig.~\ref{fig:corr}, Pearson $r{=}0.8792$).

\begin{center}
\begin{minipage}{\linewidth}
\centering
	\centering
	\includegraphics[width=0.95\columnwidth]{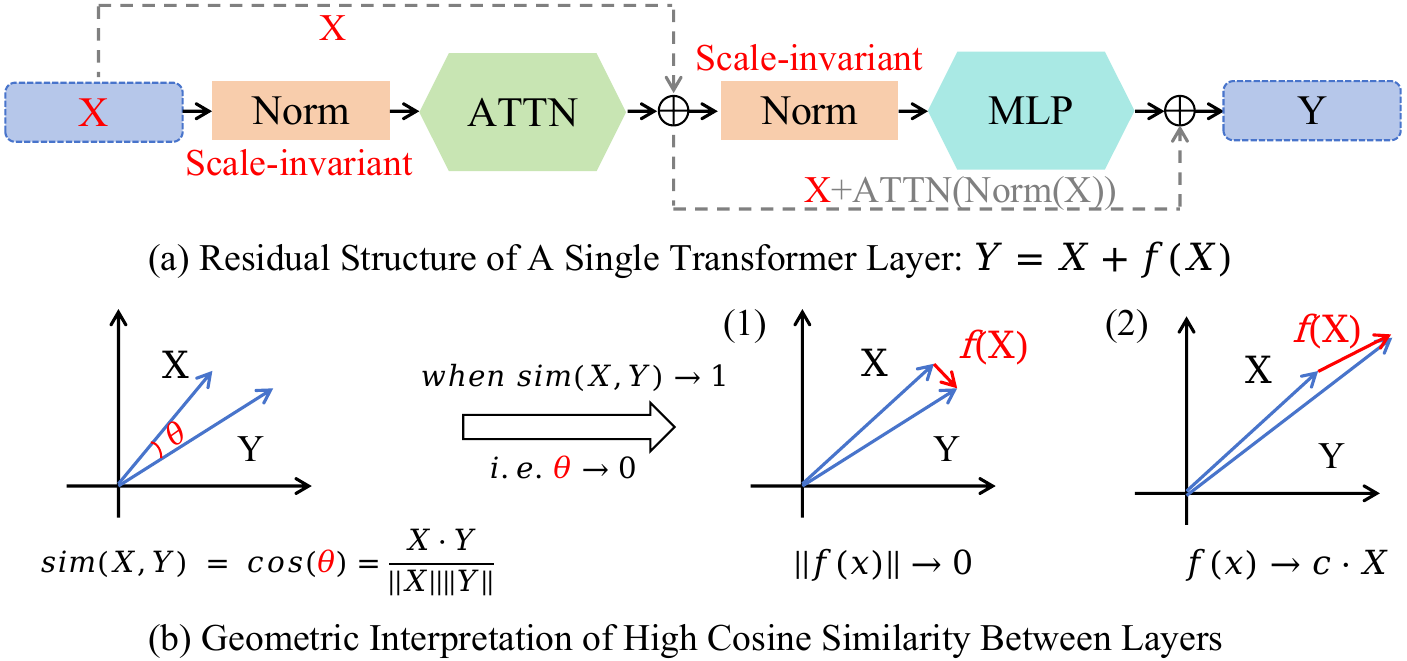}
	\captionof{figure}{Motivation. (a)~The residual structure lets the input
	$\mathbf{x}$ flow directly to the output $\mathbf{y}=\mathbf{x}+f(\mathbf{x})$;
	(b)~a high input--output cosine similarity implies $f(\mathbf{x})$ is either
	near zero or collinear with $\mathbf{x}$---both corresponding to low
	effective rank and high compressibility.}
	\label{fig:motivation}
\end{minipage}
\end{center}

\begin{center}
\begin{minipage}{\linewidth}
\centering

	\centering
	\includegraphics[width=0.8\columnwidth]{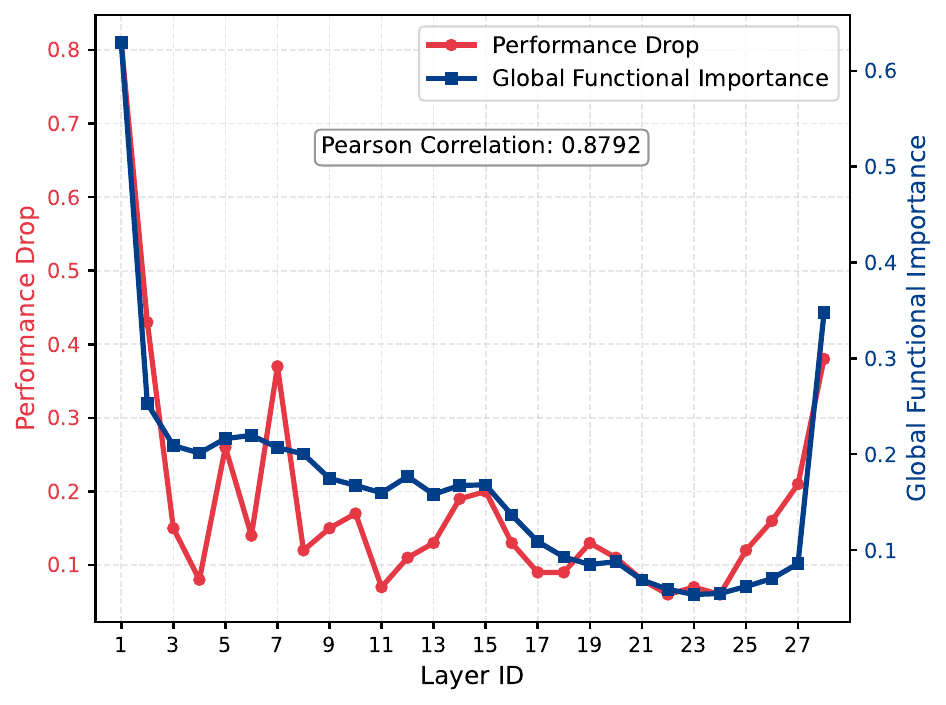}
	\captionof{figure}{Perplexity drop caused by truncating each layer at $90\%$ retained
	singular energy versus the layer functional importance $\mathcal{I}_l$ on
	Llama-3.2-3B. A strong linear correlation (Pearson $r{=}0.8792$) confirms
	$\mathcal{I}_l$ as a reliable compressibility proxy.}
	\label{fig:corr}

\end{minipage}
\end{center}

\textbf{Scoring and truncation.} Inside layer $l$, a singular triplet
$(\sigma_{l,i},\mathbf{u}_{l,i},\mathbf{v}_{l,i})$ also has a \emph{local}
structural value: the fraction of matrix energy it preserves,
$\sigma_{l,i}^2/\|\boldsymbol{\Sigma}_l\|_F^2$. S\&T combines the two views
into one score
\begin{equation}
	s_{l,i} \;=\; \mathcal{I}_l \cdot \frac{\sigma_{l,i}^2}{\lVert\boldsymbol{\Sigma}_l\rVert_F^2},
	\label{eq:score}
\end{equation}
S\&T then ranks
\emph{all} triplets of \emph{all} layers in a single pool by $s_{l,i}$. Under a target sparsity $\gamma$, the parameter budget is
$(1-\gamma)\times\#\text{params}$. We walk the sorted pool in descending order
and keep a triplet whenever its $m{+}n$ parameter cost still fits the
remaining budget---an \emph{accumulative rank preservation} that lets the
allocation adapt to heterogeneous matrix shapes, instead of applying one ratio
per layer as module-level schemes~\cite{lin2025modegpt} do. S\&T only needs
forward passes over the calibration set: no gradient is involved, allocation
takes minutes, and because it consumes only the triplet representation of any
SVD-type decomposition, it can be applied post-hoc to existing decomposers.

\textbf{Rank-preserving fine-tuning.} Adapting a decomposed model with
LoRA~\cite{hu2022lora} normally requires reconstructing the full-rank weight,
merging the LoRA delta, and re-decomposing and re-truncating to the budget,
which
discards information. RPFT instead splits the retained triplets of each matrix
into a frozen majority $\mathcal{F}$ and a small trainable subset
$\mathcal{T}$ (its 8--16 lowest-scoring retained triplets) and optimizes only
$\mathcal{T}$ in factorized form:
\begin{equation}
	\widetilde{\mathbf{W}}^{(t)}
	=\sum_{i\in\mathcal{F}}\sigma_i\,\mathbf{u}_i\mathbf{v}_i^{\top}
	+\sum_{j\in\mathcal{T}}\sigma_j^{(t)}\,\mathbf{u}_j^{(t)}
	(\mathbf{v}_j^{(t)})^{\top},
	\label{eq:rpft}
\end{equation}
where $(t)$ marks trainable quantities. Because every term stays inside the
allocated subspace, $\mathrm{rank}\big(\widetilde{\mathbf{W}}^{(t)}\big)\leq
|\mathcal{F}|+|\mathcal{T}|\leq k$ holds at every step \emph{by construction},
so the compression budget is never exceeded, re-decomposition is eliminated,
and no truncation loss is incurred.

\section{Experiments}
\label{sec:exp}

\textbf{Setup.} We evaluate on Llama-2-7B (MHA)~\cite{touvron2023llama2openfoundation}
and Llama-3.1-8B / Llama-3.2-3B (GQA)~\cite{grattafiori2024llama}. Reasoning
accuracy is measured on BoolQ, OBQA, PIQA, Winogrande, HellaSwag, ARC-E and
ARC-C with lm-eval-harness~\cite{eval-harness}; language modeling uses WikiText-2
(WT2)~\cite{Merity2016PointerSM} perplexity. Rank allocation is tested on weight-based SVD,
activation-based decompositions (AFM~\cite{yu2023compressing} and AWSVD,
LoRAP's activation-aware decomposition~\cite{li2024lorap}), and
Dobi-SVD~\cite{qinsi2025dobisvd}; LoRAP's hand-crafted
allocation~\cite{li2024lorap} serves as the main baseline on Llama-3.1-8B.
Importance scores are calibrated on 128 WikiText-2 sequences. Where
fine-tuning applies, all methods share a matched budget on a single RTX 6000
Ada: 2000 steps with batch size 16 on RedPajama (about 7 hours), while S\&T
adds roughly two minutes of calibration and no re-decomposition.

\begin{center}
	\begin{minipage}{\linewidth}
	\centering
	
		\centering
		\captionof{table}{Zero-shot WikiText-2 perplexity of manual allocation versus S\&T on Llama-3.2-3B at $40\%$ parameter retention (lower is better).}
		\label{tab:zeroshot}
		\small
		\begin{tabular}{lccc}
			\toprule
			Decomp. & Manual & S\&T (ours) & Reduction \\
			\midrule
			SVD & 762.52 & 291.48 & 61.8\% \\
			AFM & 146.63 & 86.88 & 40.8\% \\
			AWSVD & 127.73 & 55.83 & 56.3\% \\
			\bottomrule
		\end{tabular}
	
	\end{minipage}
\end{center}

\textbf{Zero-shot compression.} Table~\ref{tab:zeroshot} isolates the rank
allocation quality on Llama-3.2-3B at $40\%$ parameter retention without any
fine-tuning. Substituting S\&T for each decomposition's manual assignment
cuts perplexity by $40.8\%$--$61.8\%$ (avg.\ $52.9\%$), with the largest
gain on plain SVD---confirming that the budget, not the decomposition, limits
current pipelines.

\begin{table*}[t]
	\centering
	\caption{Average reasoning accuracy (\%) per task and WikiText-2
	perplexity on Llama-3.1-8B at $25\%$ sparsity; best in \textbf{bold},
	second best \underline{underlined}. Baseline numbers (SliceGPT~\cite{ainslie2024slicegpt},
	ShortGPT~\cite{men2024shortgptlayerslargelanguage},
	Shortened-Llama~\cite{kim2024shortened}, MoD~\cite{raposo2024mixture},
	D-LLM~\cite{jiang2024d}, SkipGPT) are reported by
	SkipGPT~\cite{zhao2025skipgpt}; LoRAP and LoRAP+UniRank were run by us
	under identical decomposition and fine-tuning budgets.}
	\label{tab:main}
	\small
	\begin{tabular}{@{}l|ccccccc|cc@{}}
			\toprule
			\multirow{2}{*}{Method} & \multicolumn{7}{c}{Reasoning (Acc. $\uparrow$)} & \multirow{2}{*}{AVG} & \multirow{2}{*}{WT2$\downarrow$} \\
			\cmidrule{2-8}
			& BoolQ & OBQA & PIQA & WinoG. & Hella. & ARC-C & ARC-E & & \\
			\midrule
			Dense & 82.14 & 44.60 & 81.07 & 77.43 & 81.89 & 57.68 & 84.81 & 72.80 & 7.33 \\
			\midrule
			SliceGPT & 72.39 & 34.40 & 66.70 & 61.56 & 56.96 & 31.48 & 50.08 & 53.37 & \textbf{9.22} \\
			Short.-Llama & 71.19 & 37.40 & 73.72 & \textbf{71.82} & 69.56 & 44.45 & 66.88 & 62.15 & 10.32 \\
			ShortGPT & 72.05 & \underline{38.40} & 73.94 & \underline{70.96} & 69.23 & 43.86 & 68.01 & 62.35 & 11.13 \\
			MoD & 50.28 & 31.60 & 64.25 & 52.41 & 50.44 & 28.24 & 37.67 & 44.98 & 34.21 \\
			D-LLM & 50.36 & 30.20 & 57.40 & 52.49 & 37.64 & 28.16 & 37.12 & 41.91 & 40.12 \\
			SkipGPT & 70.67 & 29.60 & 56.96 & 62.83 & \textbf{74.22} & \textbf{49.91} & \textbf{78.79} & 60.43 & 10.53 \\
			\midrule
			LoRAP & \underline{74.50} & 37.00 & \underline{76.06} & 65.54 & 70.33 & 44.97 & 73.23 & \underline{63.09} & 12.82 \\
			+UniRank & \textbf{76.87} & \textbf{40.40} & \textbf{78.31} & 69.90 & \underline{73.77} & \underline{47.67} & \underline{75.49} & \textbf{66.05} & \underline{10.29} \\
			\bottomrule
	\end{tabular}

\end{table*}

\textbf{Main results.} Table~\ref{tab:main} compares UniRank with
parameter- and token-removal baselines on Llama-3.1-8B at $25\%$ sparsity;
the provenance and budgets of every baseline are given in the table caption.
UniRank improves LoRAP's average reasoning accuracy by $+3.0$ points while
cutting WT2 perplexity by $2.5$, and reaches the highest accuracy among all
compressed models. Two conclusions follow: parameter-removal methods clearly dominate
token-removal ones (MoD, D-LLM, SkipGPT), and UniRank's gains over LoRAP---the
same decomposition family and fine-tuning budget---isolate the contribution of
the allocation itself.

\textbf{Plug-and-play and ablation.} Table~\ref{tab:plug} shows UniRank as a
post-hoc allocator for four decomposition methods on Llama-2-7B at $40\%$
sparsity: replacing each method's allocation with S\&T improves both
zero-shot and fine-tuned perplexity and reasoning accuracy, including a
$+1.95$ point gain over the learning-based Dobi-SVD. Table~\ref{tab:ablation} isolates the components on
Llama-3.1-8B at $25\%$ sparsity: adding local energy (S\&T$_{\mathrm{local}}$)
to a uniform baseline gives $+1.61$ points, adding the global factor gives
another $+0.97$, and RPFT contributes $+0.38$ points with a $-0.56$ perplexity
reduction on top of full S\&T with standard LoRA---each ingredient pays off,
and RPFT preserves the allocation that S\&T computed. Table~\ref{tab:group} confirms the single global pool: size-proportional
parameter- or module-level groups underperform the unified ranking
($63.44$ / $65.26$ vs.\ $66.05$ avg.\ accuracy) because groups cannot
exchange rank budget.

\begin{center}
\begin{minipage}{\linewidth}
\centering

	\centering
	\captionof{table}{UniRank as a plug-and-play rank allocator on Llama-2-7B at $40\%$
	sparsity (FT = fine-tuning, Rea. = reasoning accuracy).}
	\label{tab:plug}
	\small
	\setlength{\tabcolsep}{4pt}
	\begin{tabular}{llcc@{}c}
		\toprule
		Decomp. & Alloc. & ZS-WT2$\downarrow$ & FT-WT2$\downarrow$ & FT-Rea.$\uparrow$ \\
		\midrule
		Dense & --- & --- & 8.70 & 69.73 \\
		\midrule
		SVD & Uniform & 134.68 & 28.57 & 49.20 \\
		& +S\&T & \textbf{86.20} & \textbf{26.10} & \textbf{49.50} \\
		\midrule
		AFM & Uniform & 43.93 & 23.12 & 52.80 \\
		& +S\&T & \textbf{28.06} & \textbf{22.01} & \textbf{53.97} \\
		\midrule
		AWSVD & Heuristic & 28.05 & 23.90 & 53.67 \\
		& +S\&T & \textbf{26.75} & \textbf{22.89} & \textbf{54.02} \\
		\midrule
		Dobi & Learning & --- & 14.10 & 61.10 \\
		& +S\&T & --- & \textbf{11.74} & \textbf{63.05} \\
		\bottomrule
	\end{tabular}

\end{minipage}
\end{center}

\begin{center}
\begin{minipage}{\linewidth}
\centering

	\centering
	\captionof{table}{Component ablation on Llama-3.1-8B at $25\%$ sparsity, built from
	a uniform-allocation baseline with standard LoRA.}
	\label{tab:ablation}
	\small
	\begin{tabular}{lcc}
		\toprule
		Config. & Avg. Acc. $\uparrow$ & WT2 PPL $\downarrow$ \\
		\midrule
		Baseline & 63.09 & 12.82 \\
		+S\&T (local) & 64.70 & 12.11 \\
		+S\&T (local+global) & 65.67 & 10.85 \\
		+RPFT & \textbf{66.05} & \textbf{10.29} \\
		\bottomrule
	\end{tabular}

\end{minipage}
\end{center}

\begin{center}
\begin{minipage}{\linewidth}
\centering

	\centering
	\captionof{table}{Grouping ablation (Llama-3.1-8B, $25\%$ sparsity):
	size-proportional parameter- or module-level groups versus the unified
	global pool.}
	\label{tab:group}
	\small
	\begin{tabular}{lcc}
		\toprule
		Grouping & Avg. Acc. $\uparrow$ & WT2 PPL $\downarrow$ \\
		\midrule
		Parameter-level & 63.44 & 13.01 \\
		Module-level & 65.26 & 10.93 \\
		Unified global (ours) & \textbf{66.05} & \textbf{10.29} \\
		\bottomrule
	\end{tabular}

\end{minipage}
\end{center}

\textbf{Efficiency.} S\&T involves no gradient optimization and only a few
minutes of calibration, and it removes the re-decomposition step that LoRA
merging would otherwise require; the fine-tuning budget is shared by all
methods. Table~\ref{tab:latency} verifies that inference cost stays that of
the adopted decomposition: latency matches LoRAP at all lengths and is
$\sim$40\% lower than the dense model at 2048 tokens.

\begin{center}
\begin{minipage}{\linewidth}
\centering

	\centering
	\captionof{table}{Inference time (s) per forward pass on Llama-2-7B ($50\%$
	sparsity).}
	\label{tab:latency}
	\small
	\begin{tabular}{lccc}
		\toprule
		Seq. len & Dense & LoRAP & UniRank \\
		\midrule
		64 & 0.0157 & 0.0128 & 0.0124 \\
		128 & 0.0211 & 0.0143 & 0.0149 \\
		256 & 0.0246 & 0.0172 & 0.0177 \\
		512 & 0.0426 & 0.0258 & 0.0255 \\
		1024 & 0.0840 & 0.0489 & 0.0467 \\
		2048 & 0.1783 & 0.1066 & 0.1045 \\
		\bottomrule
	\end{tabular}

\end{minipage}
\end{center}

\section{Conclusion}
\label{sec:conclusion}
This paper presented UniRank, a unified pipeline for low-rank LLM compression.
Sorting-and-Truncation turns rank allocation into a training-free global
sorting of singular components under explicit parameter costs, guided by a
two-factor score combining local singular energy with a geometry-grounded
global functional importance; Rank-Preserving Fine-Tuning adapts the
compressed model within the allocated rank budget, eliminating
re-decomposition and truncation loss. On Llama-family models UniRank reduces
zero-shot perplexity by up to $50\%$, improves reasoning accuracy by $3.0$
points over LoRAP at $25\%$ sparsity, and boosts four SVD-based decomposers as
a plug-and-play module.

\bibliographystyle{IEEEbib}
\bibliography{unirank}

@inproceedings{li2024lorap,
	title={LoRAP: Transformer Sub-Layers Deserve Differentiated Structured Compression for Large Language Models}, 
	author={Guangyan Li and Yongqiang Tang and Wensheng Zhang},
	booktitle={International Conference on Machine Learning},
	year={2024},
}

@inproceedings{
	qinsi2025dobisvd,
	title={Dobi-{SVD}: Differentiable {SVD} for {LLM} Compression and Some New Perspectives},
	author={Wang Qinsi and Jinghan Ke and Masayoshi Tomizuka and Kurt Keutzer and Chenfeng Xu},
	booktitle={The Thirteenth International Conference on Learning Representations},
	year={2025},
}

@article{hu2022lora,
	title={Lora: Low-rank adaptation of large language models.},
	author={Hu, Edward J and Shen, Yelong and Wallis, Phillip and Allen-Zhu, Zeyuan and Li, Yuanzhi and Wang, Shean and Wang, Lu and Chen, Weizhu and others},
	journal={ICLR},
	volume={1},
	number={2},
	pages={3},
	year={2022}
}

@inproceedings{frantar2023sparsegpt,
	author = {Frantar, Elias and Alistarh, Dan},
	title = {SparseGPT: massive language models can be accurately pruned in one-shot},
	year = {2023},
	publisher = {JMLR.org},
	booktitle = {Proceedings of the 40th International Conference on Machine Learning},
	articleno = {414},
	numpages = {15},
	location = {Honolulu, Hawaii, USA},
	series = {ICML'23}
}

@inproceedings{
	ainslie2024slicegpt,
	title={Slice{GPT}: Compress Large Language Models by Deleting Rows and Columns},
	author={Saleh Ashkboos and Maximilian L. Croci and Marcelo Gennari do Nascimento and Torsten Hoefler and James Hensman},
	booktitle={The Twelfth International Conference on Learning Representations},
	year={2024},
}

@misc{men2024shortgptlayerslargelanguage,
	title={ShortGPT: Layers in Large Language Models are More Redundant Than You Expect}, 
	author={Xin Men and Mingyu Xu and Qingyu Zhang and Bingning Wang and Hongyu Lin and Yaojie Lu and Xianpei Han and Weipeng Chen},
	year={2024},
	eprint={2403.03853},
	archivePrefix={arXiv},
	primaryClass={cs.CL},
}

@inproceedings{yu2023compressing,
	title={Compressing Transformers: Features Are Low-Rank, but Weights Are Not!},
	author={Hao Yu and Jianxin Wu},
	booktitle={Proceedings of the AAAI Conference on Artificial Intelligence,},
	year={2023},
}

@misc{touvron2023llama2openfoundation,
	title={Llama 2: Open Foundation and Fine-Tuned Chat Models}, 
	author={Hugo Touvron and Louis Martin and Kevin Stone and Peter Albert and others},
	year={2023},
	eprint={2307.09288},
	archivePrefix={arXiv},
	primaryClass={cs.CL},
}

@article{grattafiori2024llama,
	title={The llama 3 herd of models},
	author={Grattafiori, Aaron and Dubey, Abhimanyu and Jauhri, Abhinav and Pandey, Abhinav and Kadian, Abhishek and Al-Dahle, Ahmad and Letman, Aiesha and Mathur, Akhil and Schelten, Alan and Vaughan, Alex and others},
	journal={arXiv preprint arXiv:2407.21783},
	year={2024}
}

@inproceedings{
	zhao2025skipgpt,
	title={Skip{GPT}: Each Token is One of a Kind},
	author={Anhao Zhao and Fanghua Ye and Yingqi Fan and Junlong Tong and Jing Xiong and Zhiwei Fei and Hui Su and Xiaoyu Shen},
	booktitle={Forty-second International Conference on Machine Learning},
	year={2025},
}

@inproceedings{
	kim2024shortened,
	title={Shortened {LL}a{MA}: A Simple Depth Pruning for Large Language Models},
	author={Bo-Kyeong Kim and Geonmin Kim and Tae-Ho Kim and Thibault Castells and Shinkook Choi and Junho Shin and Hyoung-Kyu Song},
	booktitle={ICLR 2024 Workshop on Mathematical and Empirical Understanding of Foundation Models},
	year={2024},
}

@article{raposo2024mixture,
	title={Mixture-of-depths: Dynamically allocating compute in transformer-based language models},
	author={Raposo, David and Ritter, Sam and Richards, Blake and Lillicrap, Timothy and Humphreys, Peter Conway and Santoro, Adam},
	journal={arXiv preprint arXiv:2404.02258},
	year={2024}
}

@article{jiang2024d,
	title={D-llm: A token adaptive computing resource allocation strategy for large language models},
	author={Jiang, Yikun and Wang, Huanyu and Xie, Lei and Zhao, Hanbin and Qian, Hui and Lui, John and others},
	journal={Advances in Neural Information Processing Systems},
	volume={37},
	pages={1725--1749},
	year={2024}
}

@inproceedings{Merity2016PointerSM,
	title={Pointer Sentinel Mixture Models},
	author={Stephen Merity and Caiming Xiong and James Bradbury and Richard Socher},
	booktitle={ICLR},
	year={2017},
	volume={abs/1609.07843},
}

@misc{eval-harness,
	author       = {Gao, Leo and Tow, Jonathan and Abbasi, Baber and Biderman, Stella and Black, Sid and DiPofi, Anthony and Foster, Charles and Golding, Laurence and Hsu, Jeffrey and Le Noac'h, Alain and Li, Haonan and McDonell, Kyle and Muennighoff, Niklas and Ociepa, Chris and Phang, Jason and Reynolds, Laria and Schoelkopf, Hailey and Skowron, Aviya and Sutawika, Lintang and Tang, Eric and Thite, Anish and Wang, Ben and Wang, Kevin and Zou, Andy},
	title        = {The Language Model Evaluation Harness},
	month        = 07,
	year         = 2024,
	publisher    = {Zenodo},
	version      = {v0.4.3},
	doi          = {10.5281/zenodo.12608602},
}

@inproceedings{wang2025svd,
	title={Svd-llm: Truncation-aware singular value decomposition for large language model compression},
	author={Wang, Xin and Zheng, Yu and Wan, Zhongwei and Zhang, Mi},
	booktitle={International Conference on Learning Representations},
	volume={2025},
	pages={19299--19319},
	year={2025}
}

@inproceedings{lin2025modegpt,
	title={Modegpt: Modular decomposition for large language model compression},
	author={Lin, Chi-Heng and Gao, Shangqian and Smith, James and Patel, Abhishek and Tuli, Shikhar and Shen, Yilin and Jin, Hongxia and Hsu, Yen-Chang},
	booktitle={International Conference on Learning Representations},
	volume={2025},
	pages={101355--101390},
	year={2025}
}

@inproceedings{lu2025flrc,
	title={FLRC: Fine-grained Low-Rank Compressor for Efficient LLM Inference},
	author={Lu, Yu-Chen and Chen, Chong-Yan and Chang, Chi-Chih and Hu, Yu-Fang and Wu, Kai-Chiang},
	booktitle={Proceedings of the 2025 Conference on Empirical Methods in Natural Language Processing},
	pages={14956--14966},
	year={2025}
}

@inproceedings{gao-etal-2024-adaptive,
	title = "Adaptive Rank Selections for Low-Rank Approximation of Language Models",
	author = "Gao, Shangqian  and
	Hua, Ting  and
	Hsu, Yen-Chang  and
	Shen, Yilin  and
	Jin, Hongxia",
	booktitle = "Proceedings of the 2024 Conference of the North American Chapter of the Association for Computational Linguistics: Human Language Technologies (Volume 1: Long Papers)",
	year = "2024",
	publisher = "Association for Computational Linguistics",
	pages = "227--241",
}

\end{document}